\documentclass[preprint,12pt]{elsarticle}

\usepackage{amssymb}
\usepackage{amsmath}

\usepackage{amsmath,amssymb,amsfonts}
\usepackage{algorithmic}
\usepackage{graphicx}
\usepackage{algorithm,algorithmic}
\usepackage{hyperref}
\hypersetup{hidelinks=true}
\usepackage{textcomp}

\usepackage{soul}
\usepackage{xcolor}
\sethlcolor{yellow}
\usepackage{multirow}
\usepackage{multicol}
\usepackage{booktabs}
\usepackage{amssymb}
\usepackage{pifont}
\usepackage{bbding}

\usepackage{colortbl}

\journal{Biomedical Signal Processing and Control}

\begin{document}

\begin{frontmatter}



\title{Grounded Knowledge-Enhanced Medical Vision-Language Pre-training for Chest X-Ray}

\author[label1,label2]{Qiao Deng}
\author[label3,label4]{Zhongzhen Huang}
\author[label1,label2]{Yunqi Wang}
\author[label1,label2]{Zhichuan Wang}
\author[label5]{Zhao Wang}
\author[label3,label4]{Xiaofan Zhang}
\author[label5]{Qi Dou}
\author[label6]{Yeung Yu Hui}
\author[label1,label2,label7]{Edward S. Hui\corref{cor1}}
\cortext[cor1]{Corresponding author. E-mail: edward.s.hui@gmail.com}
\affiliation[label1]{organization={Department of Imaging and Interventional Radiology, The Chinese University of Hong Kong},
            city={HKSAR},
            country={China}}
\affiliation[label2]{organization={CU Lab for AI in Radiology (CLAIR), The Chinese University of Hong Kong},
            city={HKSAR},
            country={China}}


\affiliation[label3]{organization={Shanghai Jiao Tong University},
            city={Shanghai},
            postcode={200240}, 
            country={China}}

\affiliation[label4]{organization={Shanghai AI Laboratory},
            city={Shanghai},
            postcode={200240}, 
            country={China}}

\affiliation[label5]{organization={Department of Computer Science and Engineering, The Chinese University of Hong Kong},
            city={HKSAR},
            country={China}}

\affiliation[label6]{organization={China Unicom Global Limited},
            city={HKSAR},
            country={China}}

\affiliation[label7]{organization={Department of Psychiatry, The Chinese University of Hong Kong},
            city={HKSAR},
            country={China}}
\begin{abstract}
Medical foundation models have the potential to revolutionize healthcare by providing robust and generalized representations of medical data. Medical vision-language pre-training has emerged as a promising approach for learning domain-general representations of medical image and text. Current algorithms that exploit global and local alignment between medical image and text could however be marred by redundant information in medical data. To address this issue, we propose a grounded knowledge-enhanced medical vision-language pre-training (GK-MVLP) framework for chest X-ray. In this framework, medical knowledge was grounded to the appropriate anatomical regions by using a transformer-based grounded knowledge-enhanced module for fine-grained alignment between textural features of medical knowledge and the corresponding anatomical region-level visual features. The performance of GK-MVLP was competitive with or exceeded the state of the art on downstream image understanding tasks (chest X-ray disease classification, disease localization),  generative task (report generation), and vision-language understanding task (medical visual question-answering). Our results demonstrate the advantage of incorporating grounding mechanism to remove biases and improve the alignment between chest X-ray image and radiology report.
\end{abstract}

\begin{graphicalabstract}
\includegraphics[width=\textwidth]{GA_hk_edw_bspc.png}
\end{graphicalabstract}

\begin{highlights}
\item GK-MVLP uses fine-grained visual-knowledge alignment for representation learning.
\item Knowledge prompts enhance localization and prevent irrelevant information.
\item GK-MVLP achieves SOTA in classification, localization, report generation, and VQA.
\end{highlights}


\end{frontmatter}


\section{Introduction}

\label{Introduction}

Medical foundation models could potentially revolutionize healthcare by providing robust and generalized representations of medical data, and facilitating information sharing across various tasks \cite{moor2023foundation}. These models are designed to integrate and interpret multi-modal medical data, such as electronic health records, medical images, and data from diagnostic tests. Medical vision-language pre-training (VLP) is one of the notable advancements \cite{moor2023foundation, moon2022multi} for medical foundation models by leveraging large-scale multi-modal data to learn domain-general cross-modal alignment 
during pre-training. VLP model can be subsequently generalized to various downstream tasks through task-oriented modification of model architecture and fine-tuning.

Although achieving a robust and generalizable medical foundation model is highly desired, building such a model remains highly challenging. Model training requires vast amounts of data, the access to large-scale, diverse and labeled of which is typically limited. Also, the inherent complexity and variability of multi-modal medical data impose significant challenges on the design of medical VLP, particularly in accurately and effectively aligning cross-modal features.
\begin{figure}[!t]
\centerline{\includegraphics[width=\columnwidth]{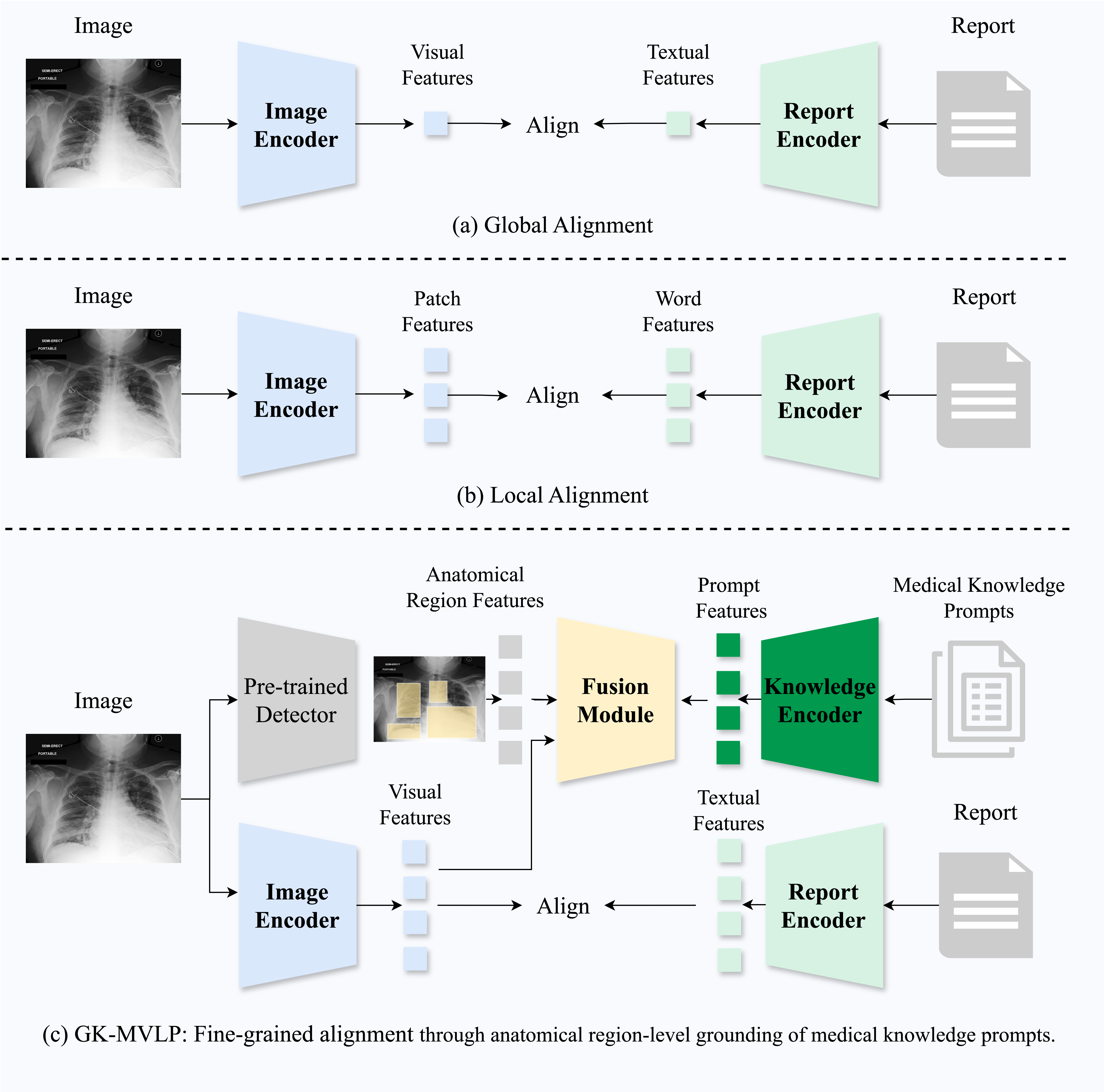}}
\caption{Comparison of different alignment methods for medical vision-language pre-training (VLP): previous models employ (a) global alignment by associating overall visual features with textual features, and (b) local alignment by connecting image patches with corresponding word features. (c) Our GK-MVLP aligns the global-local visual features (anatomical-region level) with medical knowledge features.}
\label{fig1}
\end{figure}
Recently, numerous VLP models utilize self-supervised learning to enhance the alignment between chest X-ray images and radiology reports, thereby achieving state-of-the-art performance on downstream tasks. Global and local cross-modality alignment techniques are widely used in contrastive learning. For instance, global visual and textual features have been used to align chest X-ray images with corresponding radiology reports  (see Fig. \ref{fig1}.a) \cite{zhang2022contrastive,liu2023m}. Other studies \cite{huang2021gloria, muller2022joint, boecking2022making} demonstrate that feature alignment at a fine granularity can be improved by aligning local image patches with the corresponding sentence in the report (see Fig. \ref{fig1}.b). While global and local alignment can improve granularity, these methods may fail to account for complex and subtle clinical nuances and medical terminology that require deeper contextual understanding \cite{zhang2023knowledge}. 

To address these limitations, previous studies \cite{zhang2023knowledge, wu2023medklip, lin2023towards, liu2021exploring} have incorporated posterior and prior knowledge in their multi-modal models to assist representation learning. These studies aim to improve their pre-training framework's ability to interpret and align complex medical information by incorporating the definitions of and relationships amongst the clinical entities in radiology reports. 
However, robust alignment between medical image and radiology report remains challenging as the former contains fine-grained anatomical and abnormality-specific visual features, whilst the latter could vary significantly in granularity due to differences in reporting styles and contain uncertainty due to inconclusive findings \cite{irvin2019chexpert, johnson2019mimic}. 




Considering that contrastive learning-based cross-modality alignment is semantically weak, alignment between image-level abnormality and the corresponding finding in the radiology report could be semantically inconsistent. Knowledge enhancement may unfavorably introduce medical knowledge that is not specific to training samples. To this end, we propose a grounded knowledge-enhanced medical vision-language pre-training (GK-MVLP) framework that exploits anatomical region-level grounding of the findings in radiology reports to improve the learning of domain-general representations of chest X-rays and radiology reports (see Fig. \ref{fig1}.c). To achieve this, we adopt a pre-trained detector to extract the anatomical region and pre-process the medical knowledge prompts by reformulating the clinical entities and their corresponding position information for each instance (extracted from radiology reports and well-annotated labels). We then establish the connections between abnormalities and the underlying anatomical structures using a transformer-based attention mechanism. This reduces the risk of mismatches that could occur with traditional contrastive learning-based approaches and prevents the introduction of unrelated medical knowledge that could confuse the model. 
The main contributions of this paper are: 

\begin{itemize}
    \item We propose a GK-MVLP framework that leverages fine-grained alignment between visual information and medical knowledge to assist representation learning; 

    \item Medical knowledge prompts are constructed to provide instance-level abnormality location information to prevent injecting irrelevant knowledge in the decoding stage; 

    \item Experiments show that GK-MVLP is competitive with or exceeds the performance of the state of the art on downstream image understanding tasks, namely disease classification, disease localization, downstream generative tasks, namely report generation, and downstream vision-language understanding tasks, namely medical visual question answering task. 
\end{itemize}

\section{Related Work}
\label{Related Work}

Representation learning of medical image and medical texts has been achieved using contrastive learning. ConVIRT \cite{zhang2022contrastive} employs a contrastive loss to align global representations, while GLoRIA \cite{huang2021gloria}, LovT \cite{muller2022joint} and BioViL \cite{boecking2022making} incorporate instance-level local contrastive learning on image sub-regions and corresponding radiology report sentence representations. Apart from standard global multi-modality alignment methods, a few studies exploited fine-grained representations \cite{wang2022multi,cheng2023prior, li2024anatomical}. MGCA integrates instance-wise alignment, and token-wise alignment with a bidirectional cross-attention strategy for fine-grained visual and text token matching and disease-level alignments to enhance medical visual representation learning \cite{wang2022multi}. PRIOR \cite{cheng2023prior} utilizes a prototype memory bank for sentence-wise embeddings to focus on localized visual and clinical linguistic features, and a cross-modality conditional reconstruction module that reconstructs masked images and reports for enhancing cross-modality feature interaction. Region-based representation learning has been further investigated to enhance interpretability and clinical relevance in pre-training and specialized models. The ASG framework \cite{li2024anatomical} addresses the lack of interpretability and clinical relevance in local alignment by parsing reports into triplets $\langle \text{anatomical region}, \text{finding}, \text{existence} \rangle$ for supervision, applying anatomical region-sentence alignment, and incorporating image-tag recognition to enhance semantic associations between image and report. Additionally, AdaMatch-Cyclic \cite{chen2023fine} presents an adaptive patch-word matching model that correlates adaptive image patches of abnormal regions with words in medical reports to provide interpretability for report generation.

External medical knowledge has been incorporated in medical VLP models to further enhance visual understanding \cite{wu2023medklip,zhang2023knowledge,lin2023towards}. KAD \cite{zhang2023knowledge} utilizes a knowledge graph to pre-train a knowledge encoder, optimizing the disease query network for classification tasks with paired chest X-rays and extracted entities. MedKLIP \cite{wu2023medklip} focuses on the explanation of entities and incorporates positional information into its knowledge-enhanced triplets to provide detailed supervision signals and optimize the multi-modal fusion module at the entity level. Similarly, MOTOR \cite{lin2023towards} injects structured general and specific medical knowledge into its foundation model, aiming to learn correlations among common organs and diseases, and align semantic relations at the instance level.

To further improve the fine-grained alignments between medical image and text, and reduce the underlying bias, our proposed GK-MVLP emphasizes grounding medical knowledge to the corresponding anatomical regions on chest X-rays using a transformer-based module.

\section{Methodology}
\begin{figure*}[]
\centerline{\includegraphics[width=\textwidth]{bspc_network.png}}
\caption{Illustration of the (a) pre-processing of medical knowledge prompts, and the architecture of the (b) grounded knowledge-enhanced (GK) module, where $f_{R}$ is projection head and (c) grounded knowledge-enhanced medical vision-language pre-training (GK-MVLP) framework.}
\label{fig2}
\end{figure*}
In this section, we describe the pre-processing of medical knowledge prompts, and the architecture of our proposed GK-MVLP framework, which is based on BLIP \cite{li2022blip} and consists of image encoder, report encoder, image-report encoder, report decoder and grounded knowledge-enhanced (GK) module for aligning medical knowledge prompts with the corresponding anatomical regions. The overview of the architecture of our proposed GK-MVLP framework is shown in Fig. \ref{fig2}.b and Fig. \ref{fig2}.c. Unless otherwise specified, the following formulation is concerned with a single sample for notional brevity. 


\subsection{Medical Knowledge Prompts}
A medical knowledge prompt is defined as:
\begin{align}
    \text{Prompt} = \lbrace 
\text{entity}_d, \text{position}_d, \text{exist}_d \rbrace_{d=1}^{N_{\text{entity}}},
\end{align}
where $\text{entity}_d$, $\text{position}_d$ and $\text{exist}_d$ respectively represent the name, position, and existence of the $d^{\text{th}}$ most common abnormality, and $N_{\text{entity}}$ is the number of the most common abnormalities on chest X-ray. 
A list of $N_{\text{entity}}=14$ most common abnormalities is compiled from all radiology reports (See Fig. \ref{fig2}.a for the full list). 
The position of each entity is determined from the subset of 29 different anatomical regions (see Table. \ref{tab:anatomical regions} for the full list) that are affected by the corresponding abnormality (obtained from the anatomical annotations in the chest ImaGenome dataset \cite{wu2021chest}). $\text{Prompt}$ is then constructed by querying the existence of each entity. 
During pre-training, $\text{Prompt}$ is converted into a list of sentences for entities that are present (i.e., $\text{exist}_d=\text{True}$). For instance, $\{\text{pneumonia}, \text{right mid lung zone},  \text{True}\}$ is converted to "Pneumonia is located at right mid lung zone". See Fig. \ref{fig2}.a for an overview of the pre-processing of medical knowledge prompts. 
\begin{table}[!ht]
    \caption{A list of the 29 distinct anatomical regions obtained from the Chest ImaGenome dataset. 
    \label{tab:anatomical regions}
  }
    \centering
    \begin{tabular}{ll}
    \hline
         \multicolumn{2}{c}{Anatomical Region} \\ \midrule
         right lung   &  left hemidiaphragm   \\
         right upper lung zone & trachea\\
         right mid lung zone & spine \\
         right lower lung zone& right clavicle\\
         right hilar structures & left clavicle\\
         right apical zone & aortic arch\\
         right costophrenic angle& mediastinum\\
         right hemidiaphragm & upper mediastinum\\
         left lung & superior vena cava \\ 
         left upper lung zone & cardiac silhouette\\
         left mid lung zone & cavoatrial junction\\
         left lower lung zone & right atrium\\
         left hilar structures & carina\\
         left apical zone & abdomen\\
         left costophrenic angle & {} \\
        \hline
    \end{tabular}
\end{table}




\subsection{Model Architecture}
 
\subsubsection{Backbone architecture}
The backbone architecture of GK-MVLP consists of an image encoder $E_{I}$, a report encoder $E_{T}$, an image-report encoder $E_{I \rightarrow T}$ and a report decoder $D_{T}$. 

\(E_{I}\) uses the Vision Transformer \cite{dosovitskiy2020image} to encode X-ray image \(\mathbf{I}\) as a sequence of image embeddings \(\mathbf{v} \in \mathbb{R}^{N_I}\), which are subsequently mapped to \(\mathbf{z}_{I} \in \mathbb{R}^{N_P}\) using a projection head \(f_{I}\): 

\begin{align}
    \mathbf{z}_{I} = f_{I}(\mathbf{v}=E_{I}(\mathbf{I})).
\end{align}
\(E_{T}\) employs the BERT \cite{devlin2018bert} architecture to encode the tokens of radiology report \(\mathbf{T}\) as report embeddings \(\mathbf{t} \in \mathbb{R}^{N_T}\), which are subsequently mapped to \(\mathbf{z}_{T} \in \mathbb{R}^{N_P}\) using a projection head \(f_{T}\): 
\begin{align}
    \mathbf{z}_{T} = f_{T}(\mathbf{t}=E_{T}(\mathbf{T})).
\end{align}
\(N_I\) and \(N_T\) denote the dimension of the image and report representation spaces, respectively. The image and report embeddings are projected onto the same dimension \(N_P\) for image-text contrastive (ITC) learning. 



\(E_{I \rightarrow T}\) employs a transformer encoder, which consists of self-attention layers, cross-attention layers and the feed-forward networks, to 
align visual features \(\mathbf{v}\) with report tokens \(\mathbf{T}\), and output multi-modal features \(\mathbf{z}_{I \rightarrow T}\) for the image-text matching (ITM) task:
\begin{align}
    \mathbf{z}_{I \rightarrow T} = E_{I \rightarrow T}(\mathbf{T}, \mathbf{v}, \mathbf{v}),
\end{align}
where \(\mathbf{T}, \mathbf{v}, \mathbf{v}\) are respectively the query, key and value. 

\(D_{T}\) shares the same architecture as the image-report encoder, with the key difference in the self-attention layers being replaced by causal self-attention layers. This modification enables the generation of radiology reports in an auto-regressive manner using the global-local fused features \(\mathbf{z}_{\text{fused}}\) (see \ref{sssection_gkem} for definition), and generation of report tokens \(\mathbf{Y}\) for the language modeling (LM) task.
\begin{align}
    \mathbf{Y}_{t} = D_{T}(\mathbf{Y}_{<t}, \mathbf{z}_{\text{fused}}),
\end{align}
where \(\mathbf{Y}_{<t}\) represents the previously generated report tokens up to time step \(t-1\), and \(\mathbf{Y}_{t}\) is the token generated at time step \(t\).

\subsubsection{Grounded Knowledge-Enhanced Module} \label{sssection_gkem}
The GK module consists of a pre-trained anatomical region detector \cite{tanida2023interactive}, an entity encoder \(E_{\text{entity}}\), a knowledge encoder \(E_{\text{Prompt}}\), and a fusion module $E_{\text{Fusion}}$. The GK module exploits a grounding mechanism that utilizes cross-attention to effectively integrate region-based visual features, medical knowledge prompts, and global image features, thereby enabling robust alignment of multi-modal representations as depicted in Fig. \ref{fig3}.a. Additionally, the GK module incorporates an entity classification loss $\mathcal{L}_{\mathrm{ECLS}}$, as depicted in Fig. \ref{fig3}.b, which leverages a contrastive loss function to align global image visual features with textual entity features. The complete pipeline of the GK module is illustrated in Fig. \ref{fig2}.b, demonstrating its role in grounding and aligning multi-modal representations.

\begin{figure*}[]
\centerline{\includegraphics[width=\textwidth]{bspc_fusion.png}}
\caption{Illustration of the (a) grounding mechanism and the pipeline of the (b) entity classification loss.}
\label{fig3}
\end{figure*}




The 29 anatomical region features (as defined in the chest ImaGenome\cite{wu2021chest}) of $\mathbf{I}$ is extracted and represented as $\mathbf{R} = \left [ \mathbf{r}_1,\ldots,\mathbf{r}_{29} \right ] \in  \mathbb{R}^{N_{R} \times 29}$ using a pre-trained anatomical region detector, where \(N_{R}\) represents the dimension of the anatomical region representation space. $\mathbf{R}$ is then projected to $\mathbf{z}_{R} \in \mathbb{R}^{N_\text{Prompt}}$ using a projection head \(f_{R}\):


\begin{align}
    \mathbf{z}_{R} = f_{R}(\mathbf{R}).
\label{rf}
\end{align}

The medical knowledge prompts are encoded by the frozen \(E_{\text{Prompt}}\), which shares the same architecture and parameters as \(E_T\), to produce prompt embeddings \(\mathbf{p} \in \mathbb{R}^{N_\text{Prompt}}\):

\begin{align}
    \mathbf{p} = E_{\text{Prompt}}(\text{Prompt}).
\end{align}

\(E_\text{Fusion}\) consists of two transformer decoder layers utilizing the cross-attention mechanism followed by feed-forward network. It takes anatomical region feature \(\mathbf{z}_{R}\) as a query to iteratively attend to medical knowledge prompt features \(\mathbf{p}\) for alignment and outputs locally region-based fused features \(\mathbf{z}_{\text{local}}\):
    \begin{align}
    \mathbf{z}_{\text{local}} = E_\text{Fusion}(\mathbf{z}_{R}, \mathbf{p}) = \text{FFN}(\text{Attention}(\mathbf{z}_{R}, \mathbf{p}, \mathbf{p})),
    \end{align}
where FFN represents the feedforward neural network, and $\text{Attention}$ represents the attention mechanism defined as:
    \begin{align}
        \text{Attention}(Q, K, V) = \text{softmax}\left(\frac{QK^T}{\sqrt{d_k}}\right)V,
    \end{align}
where \(\mathbf{z}_{R}\) serves as the query \(Q\), \(\mathbf{p}\) as both the key \(K\) and value \(V\). \(d_k\) represents the dimension of the keys.

The multi-modal representation \(\mathbf{z}_{\text{local}}\) is subsequently enhanced by fusing it with image embeddings \(\mathbf{v}\) derived from all patch embeddings of the image to obtain the global-local fused features \(\mathbf{z}_{\text{fused}}\):
    \begin{align}
    \mathbf{z}_{\text{fused}} = \text{FFN}(\text{Attention}(\mathbf{v}, \mathbf{z}_{\text{local}}, \mathbf{z}_{\text{local}})).
    \end{align}
\(E_\text{Fusion}\) enables the GK module to effectively fuse multi-modal information by aligning image, anatomical region, and medical knowledge prompt representations.



\(E_{\text{entity}}\) shares the same architecture and parameters as \(E_T\) and is used to encode the entity list (positive and negative) into corresponding embeddings. Specifically, the positive entities (e.g., atelectasis, infiltration, hernia) and their corresponding negative counterparts (e.g., no atelectasis, no infiltration, no hernia) are encoded as positive entity embeddings (\(\mathbf{z}_{\text{entity}}^{\text{pos}}\)) and negative entity embeddings (\(\mathbf{z}_{\text{entity}}^{\text{neg}}\)), respectively.

\begin{align}
    \mathbf{z}_{\text{entity}}^{\text{pos}} & = E_{\text{entity}}(\text{Positive Entities}), \\
    \mathbf{z}_{\text{entity}}^{\text{neg}} & = E_{\text{entity}}(\text{Negative Entities}).
\end{align}
To encourage alignment between the image embedding class tokens (\(\mathbf{v}_{\text{cls}}\)) and these entity embeddings, we apply a contrastive loss for samples that contain at least one positive entity. Let \(y_d \in \{0, 1\}\) denote the target for the \(d\)-th entity, where \(y_d = 1\) indicates the presence of the \(d\)-th entity in the image. The contrastive loss is defined as:


\begin{equation}
\mathcal{L}_{\text{ECLS}} = -\frac{1}{|N_{\text{entity}}|} \sum_{d} \log \frac{y_d \cdot \exp\left(\cos(\mathbf{v}_{\text{cls}}, \mathbf{z}_{\text{entity}, d}^{\text{pos}})/\tau\right) + (1 - y_d) \cdot \exp\left(\cos(\mathbf{v}_{\text{cls}}, \mathbf{z}_{\text{entity}, d}^{\text{neg}})/\tau\right)}{\sum_{d'} \exp\left(\cos(\mathbf{v}_{\text{cls}}, \mathbf{z}_{\text{entity}, d'}^{\text{pos}})/\tau\right) + \exp\left(\cos(\mathbf{v}_{\text{cls}}, \mathbf{z}_{\text{entity}, d'}^{\text{neg}})/\tau\right)},
\end{equation}
where \(\tau\) is the temperature parameter, and is set to 0.2.

\subsection{Pre-training Objectives}
\label{loss}
The optimization process during the pre-training of GK-MVLP involves four objectives: the ITC loss \(\mathcal{L}_{\mathrm{ITC}}\), ITM loss \(\mathcal{L}_{\mathrm{ITM}}\), LM loss \(\mathcal{L}_{\mathrm{LM}}\), and the entity-based classification loss (ECLS) \(\mathcal{L}_{\mathrm{ECLS}}\). The overall pre-training objective of GK-MVLP is given by:

\begin{align}
    \mathcal{L} = \mathcal{L}_{\mathrm{ITC}} + \lambda_1 \mathcal{L}_{\mathrm{ITM}} + \lambda_2 \mathcal{L}_{\mathrm{LM}} + \lambda_3 \mathcal{L}_{\mathrm{ECLS}},
\end{align}
where \(\lambda_1\), \(\lambda_2\), and \(\lambda_3\) are hyperparameters that balance the contributions of each loss.

\(\mathcal{L}_{\mathrm{ITC}}\) aligns the image feature \(\mathbf{z}_{I}\) with the text feature \(\mathbf{z}_{T}\) by respectively maximizing and minimizing their similarity for matching and non-matching pairs, following the methodology from \cite{li2022blip,li2021align}. It helps the model align visual features with textural features in a shared embedding space, thereby enhancing the model's understanding on the relationship between visual and textual information.

\(\mathcal{L}_{\mathrm{ITM}}\) enables fine-grained alignment between an image and its corresponding report by distinguishing whether an image-report pair is positive or negative, given the cross-modal representation \(\mathbf{z}_{I \rightarrow T}\). It ensures that the model can correctly identify whether a given report is accurately describing the corresponding image.

\(\mathcal{L}_{\mathrm{LM}}\) supervises the report generation process by the report decoder \(D_{T}\). It minimizes the cross-entropy loss for the generated report text, conditioned on the multi-modal representation \(\mathbf{z}_{\text{fused}}\). This loss ensures that the generated reports are coherent and contextually appropriate based on the multi-modal input features.

\(\mathcal{L}_{\mathrm{ECLS}}\) associates the chest X-ray image \(\mathbf{I}\) with the corresponding medical entities, encouraging the model to align image embeddings with textual entity representations. Together, these objectives enable the GK-MVLP model to effectively integrate and align multi-modal information, enhancing its ability to understand and generate medically relevant outputs.



\section{Experiment}

\subsection{Datasets} 

\subsubsection{Pre-training} 
Paired chest X-ray images and radiology reports from the MIMIC-CXR dataset \cite{johnson2019mimic} were used. The findings and impression sections of each radiology report were extracted. Medical knowledge prompts were generated based on the attributes of anatomical annotations in the Chest ImaGenome dataset \cite{wu2021chest}. In total, 166,504 paired images and reports, along with their corresponding medical knowledge prompts, were prepared for the pre-training phase.


\subsubsection{Fine-tuning} 

The National Institutes of Health (NIH) Chest X-ray dataset \cite{wang2017chestx} was used for the downstream multilabel classification task. It contains 112,120 front-view X-ray images of 30,805 unique patients with 14 common diseases labeled by the NIH. The dataset was split following the official split into 78,468 images for training, 11,219 for validation, and 22,433 for testing.

The Radiological Society of North America (RSNA) Pneumonia dataset \cite{kaggle2020pneumonia} was used for the downstream binary classification (normal or pneumonia positive) and localization tasks. The dataset was split into 70/15/15\% for the detection task, following \cite{huang2021gloria,wang2022multi}, and was split according to the official split for the classification task consisting of
25184/1500/3000 images for training, validation, and testing, respectively.

The CheXpert dataset \cite{irvin2019chexpert} was used for the downstream multilabel classification, which focused on five pathologies: atelectasis, cardiomegaly, consolidation, edema, and pleural effusion. Following \cite{zhang2022contrastive}, the official validation set was used as our test set. Additionally, following \cite{zhou2023advancing}, 5,000 images were assigned to the training set for validation. The resulting splits were 218,414 for training, 5,000 for validation, and 234 for testing.

The IU X-Ray dataset \cite{demner2016preparing} was used for the downstream report generation task. It contains 7,470 chest X-ray images and 3,955 medical reports. Following \cite{chen2022cross, chen2020generating, liu2023m, wang2022cross}, the dataset was split into 70\% for training, 10\% for validation, and 20\% for testing.

The VQA-RAD dataset \cite{lau2018dataset} was used for the downstream medical visual question-answering tasks. It contains 315 images and 3,515 question-answer pairs, covering the head, chest, and abdomen, and consists of 1,515 open-ended and 2,000 closed-ended questions, split into 3,064 question-answer pairs for training and 451 for testing. The SLAKE dataset \cite{liu2021slake} was also used for the same downstreamtask. It contains 642 radiographic images, and 14,028 question-answer pairs. The dataset was split according to the official split. 

\subsection{Implementation Details} 

The vision transformer (ViT-B/16 \cite{dosovitskiy2020image}) was adopted as the image encoder. SciBERT \cite{beltagy2019scibert} was adopted as the report encoder, knowledge encoder, and entity encoder. A 12-layer transformer encoder was adopted as the image-report encoder with hidden state dimensions set to 768. Another 12-layer transformer decoder was adopted as the report decoder. The AdamW optimizer with a weight decay of 0.05, learning rate of  \(3 \times 10^{-4}\) with a decay rate of 0.9, and a warm-up period of 3000 steps were used. The GK-MVLP was pre-trained on 8 NVIDIA A6000 GPUs, with a batch size of 32, for 20 epochs. $\lambda_1,\lambda_2$, and $\lambda_3$ were set to 1.

During downstream tasks, only the backbone network components were fine-tuned, while the GK module was exclusively utilized during pretraining. The fine-tuning procedures for each downstream task are described as follows: For disease classification, the image encoder was fine-tuned with linear classifier heads, supervised by the binary cross-entropy loss. Specifically, for the NIH dataset, the batch sizes were set to 32, 64, and 256, corresponding to 1\%, 10\%, and 100\% of the training data, respectively. Similarly, for the RSNA dataset, the batch sizes were configured as 8, 64, and 256 for 1\%, 10\%, and 100\% of the training data, respectively. For the CheXpert dataset, the batch sizes were set to 8, 32, and 128 for 1\%, 10\%, and 100\% of the training data, respectively. The training was conducted over 80 epochs with a learning rate of \(5 \times 10^{-5}\). For disease localization, the image encoder was fine-tuned using the framework proposed by \cite{wang2022multi}, with training configurations adjusted based on the proportion of training data. For 1\% of the training data, the batch size and learning rate were set to 8 and $1 \times 10^{-4}$, respectively. For 10\% of the data, the batch size was 16, and the learning rate was $5 \times 10^{-4}$. For 100\% of the data, the batch size was increased to 256 while maintaining the learning rate at $5 \times 10^{-4}$. The number of epochs was set to 50 across all configurations. In the report generation task, the image encoder and report decoder were both fine-tuned for 50 epochs with a batch size of 64 and a learning rate of \(1 \times 10^{-5}\). For the medical visual question-answering task, the image encoder, report encoder, and image-report encoder were fine-tuned on the VQA-RAD dataset based on the method described in \cite{zhan2020medical}. The batch size, number of epochs, and learning rate were set at 128, 300, and \(1 \times 10^{-4}\), respectively. For the SLAKE dataset, the image encoder, report encoder, and image-report encoder were fine-tuned based on the approach in \cite{chen2022multi} with a batch size of 8, 30 epochs, and a learning rate of \(5 \times 10^{-6}\).

\subsection{Results and Discussion}

\subsubsection{Label-Efficient Disease Classification}

GK-MVLP was fine-tuned on 1\%, 10\%, and 100\% of three widely used chest X-ray datasets, namely RSNA Pneumonia, NIH Chest X-ray and CheXpert datasets, for the downstream disease classification task. The area-under-the-curve (AUROC) scores of GK-MVLP versus state-of-the-art (SOTA) were shown in Table. \ref{cls}. GK-MVLP outperformed all CNN-based SOTA medical VLP models across different ratios of all three datasets (except the experiment using 1\% of the NIH Chest X-ray dataset). Amongst the ViT-based models, GK-MVLP ranked first or second in the majority of the experiments using the RSNA Pneumonia and CheXpert datasets. 
It was worth noting that the performance of our GK-MVLP model was competitive with or exceeded SOTA with an appreciably smaller pre-training dataset size.

\begin{table*}[!ht]
    \caption{Comparison of the area-under-the-curve (AUROC) scores for the downstream disease classification task with different training dataset sizes. \label{cls}
  }
    \centering
    \resizebox{\textwidth}{!}{
    \begin{tabular}{llccccccccc}
    \hline
        \multirow{2}{*}{Method} & \multirow{2}{*}{Pre-training Dataset\textsuperscript{a}} &\multicolumn{3}{c}{RSNA Pneumonia} & \multicolumn{3}{c}{NIH Chest X-ray}  & \multicolumn{3}{c}{CheXpert} \\
         {}&{}& 1\% & 10\% & 100\% & 1\% & 10\% & 100\% & 1\% & 10\% & 100\%  \\ \midrule
        \textit{CNN-based} & & & & & & & & & \\ 
        GLoRIA\cite{huang2021gloria} & CheXpert & 86.1 & 88.0 & 88.6 & - & - & - & 86.6 & 87.8 & 88.1   \\
        ConVIRT\cite{zhang2022contrastive} & MIMIC-CXR & 88.8 & 91.5 & 92.7 & - & - & - & 87.0 & 88.1 & 88.1  \\
        BioViL \cite{boecking2022making}& MIMIC-CXR & 88.1 & 88.4 & 89.1 & - & - & - &-& -& -  \\
        M-FLAG \cite{liu2023m} & MIMIC-CXR & - & - & - & 62.2 & 71.6 & 78.7 & 64.4 &71.4& 78.1\\
        MedKLIP \cite{wu2023medklip} & MIMIC-CXR &87.3 & 88.0 & 89.3 & 77.2 & 78.9 & 83.2 &-& -& -  \\
        KAD \cite{zhang2023knowledge}& MIMIC-CXR& 89.8&91.8& 92.5 &78.7&80.7&82.5& & & \\\midrule
        \textit{ViT-based} & & & & & & & & & \\ 
        MAE \cite{he2022masked, wang2023ecamp} & MIMIC-CXR & 84.2& 89.6&91.3 & 74.7&81.3 &\underline{85.1}&80.7&86.0&86.7 \\ 
        GLoRIA \cite{huang2021gloria,wang2023ecamp} & MIMIC-CXR& 89.7&91.2 &92.1&77.7&\underline{82.8}&85.0&86.5&87.5&87.8 \\ 
        REFERS \cite{zhou2022generalized,wang2023ecamp}& MIMIC-CXR & 89.4 & 91.6 & 92.7 & 76.7 & 80.9 & 84.7 & 87.2 & 88.1 & 88.2  \\
        MRM \cite{zhou2023advancing} & MIMIC-CXR&\bf{91.3}&\bf{92.7}&\bf{93.3}&\underline{79.4}&\bf{84.0}& \bf{85.9}& \bf{88.5}& 88.5 & 88.7\\
        ASG \cite{li2024anatomical}& MIMIC-CXR (frontal) & 88.4&89.5 &90.2 &\bf{79.5}&82.2&83.6&\underline{87.9}&\bf{89.0}& \underline{89.0} \\
        
        \rowcolor{gray!20} GK-MVLP & Chest ImaGenome & \underline{90.5}&\underline{92.1} &\underline{92.8} & 76.0 & 81.1 & 84.6 & 87.5 & \underline{88.9} & \bf{89.3}\\ \hline

    \end{tabular}    

   
    }
    \vspace{0.5em}
    \parbox{\textwidth}{\footnotesize \textsuperscript{a} The pre-training dataset size for CheXpert: 190k; MIMIC-CXR: 370k;  MIMIC-CXR (frontal): 217k; Chest ImaGenome: 160k images.} 
\end{table*}


Compared to CNN-based models (ConVIRT \cite{zhang2022contrastive}, and GLoRIA \cite{huang2021gloria}) that exploited contrastive learning, our model demonstrated improvements likely due to more fine-grained alignment. Furthermore, the results validated that our grounding mechanism—using entity positional information as prompts to align with anatomical regions and fusing these with global image features—outperformed other medical knowledge-enhanced methods such as MedKLIP \cite{wu2023medklip}, and KAD \cite{zhang2023knowledge}.

Compared to ViT-based models, the performance of our GK-MVLP was varied. In particular, MRM outperformed our model in a number of experiments, partly due to their masked record modeling, which reconstructed masked image patches and masked report tokens. This allowed the model to learn informative features from the larger MIMIC-CXR dataset. 
However, the MRM method may have overfitted to the specific characteristics of the pretraining dataset, making it less adaptable to the unique features and variability present in CheXpert. ASG associated specific anatomical regions with textual descriptions, enhancing semantic alignment between images and text. However, its focus on local alignment limited its ability to model global contextual relationships across regions. In contrast, our GK-MVLP employed a grounding mechanism to effectively capture both locally region-based and global context, leading to superior performance in most experiments compared to ASG.

\subsubsection{Label-Efficient Disease Localization}

GK-MVLP was fine-tuned on the RSNA Pneumonia dataset for the downstream disease localization task. Table. \ref{tab:detection} showed the mean average precision (mAP) of different algorithms across different sizes of the training RSNA Pneumonia dataset. 
GK-MVLP outperformed SOTA across all dataset sizes. 

Our GK-MVLP outperformed methods that exploited cross-modal alignment, such as PRIOR and MGCA, likely suggesting that our fusion module could more effectively integrate multimodal features of varying granularity (chest X-ray, anatomical region-level, and knowledge prompt text features). While MRM was better than our model in the classification task, the performance of our GK-MVLP exceeded that of MRM in the disease localization task. Taken together, the use of masked record modeling in MRM may have likely enhanced global comprehension for classification but failed to effectively capture the nuanced, localized features essential for precise disease localization.

\begin{table}[!ht]
    \caption{Comparisons of the mean average precision (\%) for the downstream disease localization task with different sizes of the training RSNA Pneumonia dataset. \label{tab:detection}
  }
    \centering
    \begin{tabular}{llccc}
    \hline
        \multirow{2}{*}{Method} & \multirow{2}{*}{Pre-training Dataset\textsuperscript{a}} &\multicolumn{3}{c}{RSNA Pneumonia} \\
        {} & {} &1\% & 10\% & 100\%  \\  \midrule
        \textit{CNN-based} & & & \\ 
        ConVIRT\cite{zhang2022contrastive} & MIMIC-CXR & 8.2& 15.6 & 17.9\\ 
        GLoRIA \cite{huang2021gloria},\cite{wang2022multi}& MIMIC-CXR & 11.6 & 16.1 & 24.8\\
        PRIOR \cite{cheng2023prior}& MIMIC-CXR (frontal) & 0.2&19.6 & 22.2\\
        MGCA\cite{wang2022multi} &MIMIC-CXR (frontal)& \underline{12.9}& 16.8& 24.9\\ \midrule
        \textit{ViT-based} & & &  \\
        MGCA \cite{wang2022multi}, \cite{wang2023ecamp}& MIMIC-CXR (frontal)& 8.9&19.2 &26.3 \\ 
        MRM \cite{zhou2023advancing}, \cite{wang2023ecamp}& MIMIC-CXR& 11.5& \underline{20.3}&\underline{27.1} \\
        \rowcolor{gray!20} GK-MVLP & Chest ImaGenome & \bf{21.0} & \bf{26.2} &  \bf{31.2} \\ \hline
    \end{tabular}    
    \vspace{0.5em}
    \parbox{\linewidth}{\footnotesize \textsuperscript{a} The pre-training dataset size for MIMIC-CXR (frontal) for PRIOR: 182k; MIMIC-CXR (frontal) for MGCA: 217k.}    
\end{table}

\subsubsection{Report Generation}

GK-MVLP was fine-tuned on the IU X-Ray dataset for the downstream report generation task. The natural language processing metrics, namely $\text{BLEU}_4$, $\text{METEOR}$, $\text{ROUGE}_L$, and $\text{CIDEr}$ scores, of GK-MVLP versus SOTA were shown in Table. \ref{tab:report generation}. Of note is that our GK-MVLP model was not compared against existing methods on MIMIC-CXR, as this dataset was used for pretraining with the same objective as the LM task. Our GK-MVLP model outperformed SOTA on all metrics, except ROUGE$_L$.

Our GK-MVLP outperformed SOTA specifically designed for radiology report generation across most evaluation metrics with simple fine-tuning. Notably, our model achieved a significantly higher CIDEr score, which rewarded the generation of relevant and diverse content by comparing the similarity of generated reports to a set of reference reports. This superior performance was likely attributed to the benefits of pre-training, which provided rich feature representations and improved language modeling, similar to MOTOR. AdaMatch-Cyclic struggled with limited feature fusion and a lack of comprehension of the global context. In contrast, our GK-MVLP addressed these limitations by supervising the global image CLS token with an entity loss and fusing region-based features with global context. These fused features were then directly fed into the report decoder, ensuring the generation of more coherent, accurate, and contextually rich reports. This approach rendered our model significantly more effective than AdaMatch-Cyclic for report generation tasks.

We compared our model with two pre-trained foundation models utilizing chest X-ray domain data, focusing on vision-language tasks. These results demonstrated that our medical knowledge injection strategy, which provided instance-level abnormality location information to prevent injecting irrelevant knowledge during decoding, outperformed the generalist model MOTOR. 
\begin{table}[!ht]
    \caption{Comparisons of the natural language processing metrics for the downstream report generation task.
  } \label{tab:report generation}
    \centering
    \resizebox{\linewidth}{!}{
    \begin{tabular}{llcccc}
    \hline
        Method & BLEU$_4$ & METEOR & ROUGE$_L$ & CIDEr  \\ \midrule  
        \textit{Specialist} & & & \\
        Con-Trans \cite{alfarghaly2021automated} & 0.111 & 0.164 & 0.289 & 0.257 \\
        SentSAT+KG \cite{zhang2020radiology} &0.147 & - & 0.367 & 0.304 \\
        PPKED \cite{liu2021exploring} &\underline{0.168} & 0.190 & \bf{0.376} & 0.351 \\ 
        R2GenCMN \cite{chen2022cross} &0.165 & 0.187 & \underline{0.371} & 0.398 \\
        AdaMatch-Cyclic \cite{chen2023fine}  &  0.145 & 0.162&  0.366&\\ \midrule
        \textit{Generalist}\textsuperscript{a} && & & \\ 
        MedViLL \cite{moon2022multi} &0.049 & - & -& - \\
        MOTOR\cite{lin2023towards} & 0.156 & \underline{0.193} & 0.314 & \underline{0.699} \\ 
        \rowcolor{gray!20} GK-MVLP &\bf{0.169} & \bf{0.198} & 0.322 & \bf{0.713} \\ \hline
    \end{tabular}
    }
    \vspace{0.5em}
    \parbox{\linewidth}{\footnotesize \textsuperscript{a} The pre-training dataset for MedViLL: MIMIC-CXR (370k images); MOTOR: MIMIC-CXR (AP; 89k images).}    
\end{table}

\subsubsection{Medical Visual Question-Answering}

GK-MVLP was fine-tuned on the VQA-RAD and SLAKE datasets for the downstream medical VQA task. Table. \ref{tab:med-vqa} summarized the mean accuracy in open and closed questions and the overall performance for each dataset of different. Our GK-MVLP model demonstrated superior performance across both datasets against SOTA specialist and generalist models. It achieved the highest overall accuracies of 72.5\% on VQA-RAD and 82.5\% on SLAKE. On the VQA-RAD dataset, it attained near-top performance with 77.9\% for closed questions and 64.2\% for open questions. On the SLAKE dataset, GK-MVLP achieved the highest accuracy for closed (86.1\%) questions.

Specialist models like QCR+TCR and CPRD-BAN performed well on closed-ended questions due to their focused training, which allowed them to excel at narrowly defined tasks (i.e. “yes/no” or other limited choices), but they struggled with open-ended questions that required a broader understanding and adaptability. Generalist models like MedViLL and MOTOR showed promise but still fell short of our GK-MVLP's superior accuracy, which underscored the effectiveness of our broad and diverse pre-training technique in developing a robust understanding of medical data for VQA tasks.
\begin{table}[!ht]
    \caption{Comparisons of the mean accuracy (\%) for the downstream medical visual question-and-answering tasks. 
  } \label{tab:med-vqa}
    \centering
    \resizebox{\linewidth}{!}{%
    \begin{tabular}{llcccccc}
    \hline
        \multirow{2}{*}{Method}&  \multicolumn{3}{c}{VQA-RAD} & \multicolumn{3}{c}{SLAKE}\\ 
         & Open & Closed & Overall & Open & Closed & Overall \\ \midrule
        \textit{Specialist} & & \\
        MFB \cite{yu2017multi} & 14.5& 74.3&50.6&72.2& 75.0&73.3  \\
        SAN \cite{yang2016stacked} & 31.3 & 69.5 & 54.3 & 74.0& 79.1&76.0    \\
        BAN \cite{kim2018bilinear} &37.4 & 72.1 & 58.3 & 74.6& 79.1 &76.3\\
        MEVF-SAN \cite{nguyen2019overcoming} & 49.2& 73.9& 64.1& 75.3& 78.4& 76.5\\
        MEVF-BAN\cite{nguyen2019overcoming} &49.2 & 77.2 & 66.1 & 77.8& 79.8& 78.6\\ 
        CPRD-BAN \cite{liu2021contrastive} &52.5&\underline{77.9} &67.8&79.5&83.4& 81.1 \\
        QCR+TCR \cite{zhan2020medical}&60.0 & \bf{79.3} & \underline{71.6} & -&- &- \\ \midrule
        \textit{Generalist}\textsuperscript{a} & &  \\ 
        
        MedViLL \cite{moon2022multi} & 59.5 & 77.7 & - &- & -& -\\
        MOTOR \cite{lin2023towards} & \bf{64.8} & 74.6 & 70.7 & \bf{81.1}& \underline{84.1} & \underline{82.3}  \\ 
        \rowcolor{gray!20} GK-MVLP & \underline{64.2} & \underline{77.9} & \bf{72.5} & \underline{80.2}& \bf{86.1}&\bf{82.5} \\ \hline
    \end{tabular}
    }
    \parbox{\linewidth}{\footnotesize \textsuperscript{a} The pre-training dataset for MedViLL: MIMIC-CXR (370k images); MOTOR: MIMIC-CXR (AP; 89k images).} 
\end{table}
\subsubsection{Ablation Study}

An ablation study was conducted to evaluate the contributions of the grounding mechanism and entity-based classification loss in our GK-MVLP framework to the performance of the downstream report generation and disease classification tasks. The results were summarized in Table. \ref{Ablation2}. The baseline model without these two key components of GK-MVLP was the vanilla BLIP \cite{li2022blip}. 


As compared to the baseline vanilla BLIP, the introduction of the grounding mechanism, defined as the fusion of anatomical region features and medical prompt features during pre-training, was further investigated in our ablation study by exploring two approaches: a simple concatenation of these features and the use of a cross-attention (CA) mechanism. While both methods slightly improved $\text{BLEU}_4$ to 0.162 for report generation and showed partial improvements in classification (e.g., on CheXpert and RSNA datasets), cross-attention performed slightly worse than concatenation in some classification tasks, likely due to its increased model complexity, which may have led to overfitting on smaller or imbalanced datasets, whereas concatenation provided a simpler, more stable representation.

The inclusion of the entity-based classification loss ($\mathcal{L}_{\mathrm{ECLS}}$) strengthened the alignment between visual and textual features by supervising the model to recognize key clinical entities. This led to improvements in both tasks when combined with the grounding mechanism, achieving the best performance in report generation and classification experiments. Specifically, when paired with cross-attention, it achieved the highest scores on the RSNA dataset (90.5\% for 1\%, 92.1\% for 10\%, and 92.8\% for 100\%), the 100\% data splits of CheXpert (89.3\%) and NIH (84.6\%) datasets for classification, and a $\text{BLEU}_4$ score of 0.169 for report generation, highlighting its effectiveness across both tasks. However, concatenation with $\mathcal{L}_{\mathrm{ECLS}}$ sometimes outperformed cross-attention in classification, possibly due to its simplicity and better generalization on imbalanced datasets but underperformed in report generation because it lacked the ability to dynamically model complex feature relationships necessary for producing coherent and detailed reports.

\begin{table*}[]
    \caption{
    Ablation study of the proposed GK-MVLP framework, evaluated on downstream report generation and disease classification tasks.\label{Ablation2}
  }
    \centering
    \resizebox{\textwidth}{!}{
    \begin{tabular}{c|cc|ccc|ccc|ccc|ccc}
    \hline
     \multirow{2}{*}{$\mathcal{L}_{\mathrm{ECLS}}$} & \multicolumn{2}{c|}{Grounding} & \multicolumn{3}{c|}{IU-Xray} &   \multicolumn{3}{c|}{NIH (AUROC)}& \multicolumn{3}{c|}{CheXpert (AUROC)} & \multicolumn{3}{c}{RSNA (AUROC)} \\ 
     
     & Concat & CA& $\text{BLEU}_4$ & METEOR & $\text{ROUGE}_L$ &   1\%& 10\%& 100\%&1\%& 10\%& 100\% &1\%& 10\%& 100\% \\ \midrule
     -& -&- &0.159 & 0.196 & 0.321& 75.7& 81.0 & \underline{84.4}& 86.0 & 88.3 & \underline{89.2} &90.2&91.8&92.5\\ 
    \Checkmark & -&- & \underline{0.163} & 0.197 & 0.317 & 74.6& \underline{81.1}&\underline{84.4}& 86.7& 88.6&  88.6&89.5& 91.7&\underline{92.7} \\
    -& \Checkmark&-&0.162 & 0.194 & 0.321& 74.9& 80.8 & 84.3 & 87.4 & 88.8& 88.8 & 90.1 & \underline{92.0} & \bf{92.8}\\

    -& - & \Checkmark& 0.158 & \bf{0.199} &\bf{0.323}& 74.9 &80.8 & 84.3 & 86.4 & 88.4 & 89.0 & 90.1& 91.8&92.5\\
    
    \Checkmark & \Checkmark& -&  0.158 & 0.194 & 0.317 & \bf{76.3} & \bf{81.6} & \bf{84.6} & \bf{87.8}&\bf{89.2}&88.6& \underline{90.4}& \underline{92.0} & \underline{92.7}\\
    \rowcolor{gray!20} \Checkmark & -& \Checkmark & \bf{0.169} & \underline{0.198} & \underline{0.322}& \underline{76.0} & \underline{81.1} & \bf{84.6} & \underline{87.5} & \underline{88.9} & \bf{89.3} & \bf{90.5} & \bf{92.1} & \bf{92.8}\\ \bottomrule
    \end{tabular}    
    }
\end{table*}


    

\subsubsection{Limitation}

The major limitations of our current study are two-fold.  Our experiments were performed using a limited amount (160k) of well-annotated image-report pairs. On the other hand, our model ought to be extended to other imaging modalities, such as CT, MRI, and ultrasound to further improve model generalizability. In future studies, we will investigate the ability of large language models to provide useful information for processing medical images from different imaging modalities. 

\section{Conclusion}

We have presented a grounded knowledge-enhanced medical vision-language pre-training framework that served as a foundation model for improving the learning of domain-general representations of chest X-ray and radiology reports. Our results showed that grounding medical knowledge with the appropriate anatomical regions permitted performance gain in various chest X-ray tasks. Specifically, our foundation model enhanced the framework's accuracy in disease classification, disease localization, report generation, and medical visual question answering, while also demonstrating strong generalizability across diverse medical imaging and vision-language tasks. This underscored the effectiveness of combining a robust foundation model with fine-grained alignment of visual and textual features. 





\bibliographystyle{elsarticle-num} 
\bibliography{paper}

\end{document}